\documentclass[a4paper, conference]{IEEEtran}

\usepackage{graphicx}

\ifCLASSINFOpdf
\else
\fi
\begin{document}
%
\title{Table Caption Generation in Scholarly Documents Leveraging Pre-trained Language Models}

\author{\IEEEauthorblockN{Junjie H. Xu}
\IEEEauthorblockA{\textit{University of Tsukuba}\\
Ibaraki, Japan\\
s2021705@s.tsukuba.ac.jp}
\and
\IEEEauthorblockN{Kohei Shinden}
\IEEEauthorblockA{\textit{University of Tsukuba}\\
Ibaraki, Japan\\
kohei.shinden@gmail.com}
\and
\IEEEauthorblockN{Makoto P. Kato}
\IEEEauthorblockA{\textit{University of Tsukuba}\\
Ibaraki, Japan\\
mpkato@acm.org
}}



\maketitle

\begin{abstract}

This paper addresses the problem of generating table captions for scholarly documents,
which often require additional information outside the table.
To this end, we propose a method of retrieving relevant sentences from the paper body,
and feeding the table content as well as the retrieved sentences 
into pre-trained language models (e.g. T5 and GPT-2) for generating table captions.
The contributions of this paper are: (1) discussion on the challenges in table captioning for scholarly documents; (2) development of a dataset \textit{DocBank-TB}, which is publicly available; and (3) comparison of caption generation methods for scholarly documents with different strategies to retrieve relevant sentences from the paper body.
Our experimental results showed that T5 is the better generation model for this task, as it outperformed GPT-2 in BLEU and METEOR implying that the generated text are clearer and more precise. 
Moreover, inputting relevant sentences matching the row header or whole table is effective.

\end{abstract}



%
\IEEEpeerreviewmaketitle

\section{Introduction}
Tables are important components for understanding scholarly documents 
when researchers need to describe details in a structural way, 
especially quantitative details~\cite{Table}. 
Comparing to tables from other sources, we notice caption generation for a table in a scholarly document is a slightly intensive problem, as a table in a scholarly document is (1) both semi-structured and knowledgeably informative, (2) requires not only information from the table but also additional information outside the table that is related to the table to understand it.
An appropriate table caption should be given to a table
so that readers can easily grasp the overview of the table or 
be prepared to understand the table content precisely.
However, there are tables with an inappropriate, uninformative, or poorly written caption.
While some studies aim to automatically generate an appropriate explanation for the table,
their application has been limited to {\it self-contained} tables,
which do not require information other than the table itself for generating appropriate captions~\cite{basket,bao2018table}. 

We address the problem of generating table captions for scholarly documents, which often require additional information outside the table. Thus, we propose a method of retrieving relevant sentences from the paper body, and feeding the table content as well as the retrieved sentences into pre-trained language models for generating table captions.

The contributions of this paper are: 

\begin{itemize}
    \item Discussion on the challenges in table captioning for scholarly documents.
    \item Development of a dataset \textit{DocBank-TB} contains of both tabular data and textual data in the body text that suits our table caption generation task, which is publicly available\footnote{https://github.com/junj2ejj/GCCE2021\_materials}.
    \item Comparison of caption generation methods for scholarly documents with different strategies to retrieve relevant sentences from the paper body.
\end{itemize}



\section{Related Work}

Natural language generation for table or data, namely automatically describing the content of a table is a fundamental problem in artificial intelligence that connects Natural Language Processing (NLP) and knowledge submerged in a structure. 
Advances in encoder-decoder models based on neural networks for data-to-text generation for tables shown some promising in generating descriptive text from table as well as structured data lead considerable attempts using neural-based encoder-decoder generative models~\cite{puduppully2019data, iso2019select, puduppully2021plan, rebuffel2020hierarchical}. Though the score of metric scores increases on the existing domain-specific dataset~\cite{basket} by introducing more advanced inferences, inferences designed by these proposed method are restricted to domain-specific content as the dataset consists similar tables with limited schemas.

Recently, unsupervised pre-training of large neural models has brought revolutionary changes to NLP~\cite{BERT} including Natural Language Generation (NLG)~\cite{T5, GPT2} as it enable the possibility leveraging the knowledge learned from large scale of data without supervision in the pre-trained session of these models. In NLG task for open-domain tables is that to generate text for tables from various of schema. Chen et al.~\cite{chen2020} proposed a numerical reasoning approach by combining annotated sentences to entail the logical fact underlying in the table, Suadaa et al.~\cite{suadaa2021} uses the representations consisting richer data of table itself such as captions, row headers, column headers, cell values, and metrics.  Difference from the aforementioned approach, our research focus on effective retrieving the relevant information outside the table as well as concerning different parts of the table, and see the caption as the reference data instead of using additional manual annotated data. 



\begin{figure*}[htbp]
  \centering
  \includegraphics[width=\linewidth]{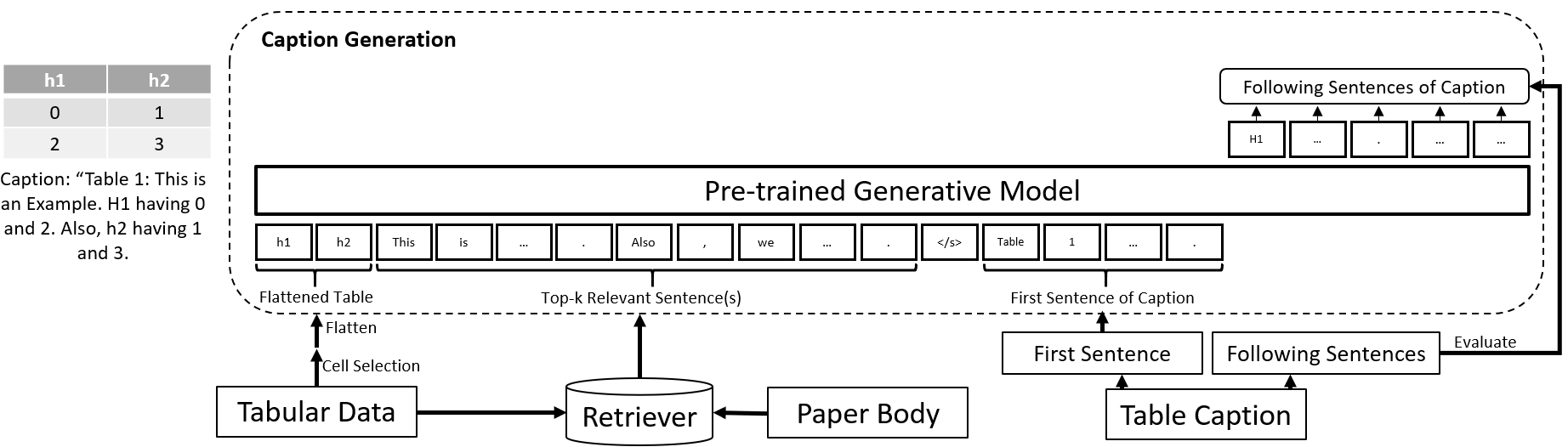}
  \caption{Illustration of the proposed method. This example was shown by using a simple table that takes row header $T_{r.h}$ as tabular data, retrieving top-2 sentences as relevant sentences using BM25 retrieval function to make the tabular and textual data representation as input to T5 model for caption generation with our proposed method.}
  \label{fig:method}
\end{figure*}

\section{Dataset Construction}
Our task requires tabular data and sentences from the paper body to generate a table caption. Since there is no public dataset for our task, we processed \textit{DocBank}~\cite{DocBank} which includes weak supervised token-level annotations of 500,000 pages (19,638 pages having semantic unit of table) in scholarly documents to develop our dataset called \textit{DocBank-TB}. In DocBank, each line represents the information of each semantic unit, contains a token of the semantic unit, bounding box, color and font. The semantic units in a page are arranged from left to right and top to bottom according to the position of bounding box existing a semantic unit. We first manually examined this dataset both layout as well as the textual information to ensure whether DocBank had been correctly extracted or not, we examine 19,638 pages and we found that (1) it is hard to determine the simple rules for arranging the sentences, the table, and the caption of the table sequentially a two-column formatted scholarly document automatically using their dataset without using advanced detection methods that out of scope of this work, (2) it is also difficult to assign caption correctly to a table when there are multiple table on the same page like a human could easily do, especially when they are very close in positional, (3) there is no a complete sentence in some pages, which means we could not retrieve the relevant sentences from such page. For reason listed above, we re-preprocessed data in the DocBank dataset follows the filtering criteria given belows:

\begin{enumerate}
    \item In one-column format,
    \item Have only one table in the page,
    \item Have a complete sentence. In practice, we select the page have three sentences or more, as it is unable to check whether there are any words before the beginning and the end of a sentence if such sentence are not between two sentences from information in a page.
\end{enumerate}

We randomly sampled pages from DocBank dataset that having table and extract the data file of those pages meet the filtering criteria described above. As a result, 500 pages was extracted follows this pipeline and used for the base of the our Dataset DocBank-TB. We processed the body text, caption and table and parsed semantic units sequencially to sentences of the body text, the caption and the rows of the table in JSON format.

In our experiment, we further derived and exploited 207 tables from those pages that have captions with more than two sentences to verify our approach. 

\section{Methods}

The goal of our table captioning task is that, given a scholarly document and a table in it, 
we generate a caption providing a better understanding of the table. An overview of our proposed method is shown in Fig. \ref{fig:method}. Our approach is to expand the first sentence of the caption with sequence-to-sequence pre-trained model T5 or GPT2 which is capable to expand the sentence to predict the next sentence of the given first sentence of caption as well as the following ones while referring to the relevant sentences and tabular data of the table.
In this work, we generate caption using the input combined with three parts: tabular data, textual data of relevant sentences from the body text and the first sentence of the caption. 
First, we extract tabular data from the table and the textual data of relevant sentences from the document. 
Then, we use the first sentence of the caption to generates the sentences that might appear after the inputted first sentence of the caption as the extension of the first sentence of the caption. 
In this section we describe the details of our proposed method.

\subsection{Tabular and Textual Data Representation}
\textbf{Table Representation as a Text Sequence}
Pre-trained transformer models used in this work are originally designed for natural language tasks, which takes word sequences as input. To apply these models taking table as input, in this work, we follow previous work~\cite{tapas} and flatten the table cells of the entire table to sequences and concatenate words in all of the cells sequentially into a sequence of words.

\begin{table*}[htbp]
\label{tab:example}
\caption{Examples of generated text using concatenation of tabular data, sentence exactly matched by table indicators (``Author''), and first sentence of caption.}
\begin{center}
\begin{tabular}{p{6em}|l|p{43em}}
 Model & Input Numeral data & Generated Text \\
 & from tabular data? & \\
 \hline
 Reference (derived from \cite{ref}) & - & The geometry of the air showers is fixed to a zenith angle of 50 degree coming from south. Each cell shows the mean of at least 20 air showers simulated with the same settings but different random seeds. The uncertainties shown are the uncertainty of the mean, and the standard deviation is shown in brackets.\\
\hline
GPT2 & No & level normalized refractivity radiation energy sea MeV at . Table 6. The table shows the normalized radiation energies for different refractivities of the atmosphere. For each atmosphere, the values of the mean of this table are the normalized energies.  The value is the standard deviation (SPD) of the measured \\
\hline
GPT2 & Yes & refractivity at sea level normalized radiation energy 2.04 × 10--4 (-30\%) 9.84 ± 0.04 (0.17) MeV (-10.3\%) 2.63 × 10--4 (-10\%) 10.63 ± 0.04 (0.18) MeV (-3.1\%) 2.77 × 10--4 (-5\%) 10.79 ± 0.05 (0.20) MeV (-1.6\%) 2.92 × 10--4 10.97 ± 0.03 (0.21) MeV 3.07 × 10--4 (+5\%) 11.12 ± 0.05 (0.21) MeV (+1.4\%) 3.21 × 10--4 (+10\%) 11.29 ± 0.04 (0.19) MeV \\
\hline
T5 & No & level normalized refractivity radiation energy sea MeV at Table 6. The table shows the normalized radiation energies for different refractivities of the atmosphere.\\
\hline
T5 & Yes & 0.21\%  0.01 (0.18)  0.05 (0.13)  0.007 (0.24)  0.001 (0.002)  0.17\%  0.13\%  0.03 (0.19)  0.07 (0.22)  10 \\

\end{tabular}

\end{center}
\end{table*}  

\textbf{Relevant Sentences Retriever}
BM25~\cite{BM25} is a retrieval function originally designed to rank a set of documents based on the term match between each document and the given query. We use the BM25 function to calculate the relevance score of each sentence in the paper body to a sequence of words from every cells of a given table, and then retrieve the top-N sentences with the highest BM25 relevance score.

\subsection{Pre-trained Models for Natural Language Generation}
We use two SOTA generative pre-trained models, \textbf{Text-to-Text Transfer Transformer (T5)}~\cite{T5} and \textbf{GPT-2}~\cite{GPT2} attempting to obtain the learned information or language representation as these models were pre-trained on large-scale of English web corpus. In terms of implementation, to preprocess data for T5, we concatenate the table, relevant sentences, the special separate token of T5 \textless/s\textgreater, and the first sentence of the caption as input to predict the following tokens. For GPT-2, since the GPT-2 model has no special separate token, we concatenate the table, relevant sentences, and the first sentence of the caption as input to predict the following tokens.

\begin{table*}[htbp]
\label{tab:result}
\caption{Evaluation results of two automatic generative models with different retrieval methods.}
\begin{center}
\begin{tabular}{ll|cccccccccccccc}
\multicolumn{2}{l|}{Retrieval Method} & None &
\multicolumn{3}{c}{Top-1 BM25} &
\multicolumn{3}{c}{Top-2 BM25} & 
\multicolumn{3}{c}{Top-3 BM25} &
Author \\
\hline
\multicolumn{2}{l|}{Table Structure} & - & $T_{r.h}$ & $T_{r.o}$ & $T_{r.w}$ &  $T_{r.h}$ & $T_{r.o}$ & $T_{r.w}$ &  $T_{r.h}$ & $T_{r.o}$ & $T_{r.w}$  & -        \\
 \hline
 Model & Metric \\
 \hline
T5 & BLEU & \textbf{8.29} & 7.80   &  7.50  &  7.40 & 7.65 & 7.05 & 7.94 & 7.02 & 6.90 & 7.00 & 7.95 \\
T5 & ROUGE-1 & 0.080 &  0.088   &  0.073  &  0.083  &  0.076 & 0.076 & 0.090  & 0.079 & 0.086 & 0.084 &  0.086   \\
T5 & ROUGE-2 & 0.009 & 0.011 & 0.009  & 0.010 & 0.009 & 0.010    & 0.016 & 0.011 & 0.012 & 0.013 &  0.015     \\
T5 & ROUGE-L & 0.070 & 0.076    &   0.063  & 0.072  & 0.065  & 0.065     & 0.079  & 0.069 & 0.070 & 0.072 & 0.070      \\
T5 & METEOR & 0.141 & \textbf{0.147}    &   0.137  & 0.136  & 0.136  & 0.136     & 0.139  & 0.132 & 0.129 & 0.127 & 0.146      \\
\hline
GPT-2 & BLEU & 5.33 & 6.92 & 6.79 & 6.99 & 6.14 & 5.91 & 5.91 & 5.33 & 5.20 & 5.19 & 6.62\\
GPT-2 & ROUGE-1 & 0.131 & 0.134   & 0.131 & \textbf{0.138} & 0.135 & 0.131 & 0.137 & 0.131 & 0.129 & 0.132 & 0.133        \\
GPT-2 & ROUGE-2 & \textbf{0.028} & 0.023    & 0.021 & 0.025 & 0.027 & 0.023 & 0.027 & \textbf{0.028} & 0.026 & 0.027 & 0.024        \\
GPT-2 & ROUGE-L & 0.093 & 0.100    & 0.094 & \textbf{0.101} & 0.097 & 0.092 & 0.097 & 0.093 & 0.089 & 0.092 & 0.099        \\
GPT-2 & METEOR & 0.058 & 0.134    & 0.131 & 0.138 & 0.135 & 0.131 & 0.137 & 0.131 & 0.129 & 0.132 & 0.089       \\
\end{tabular}

\end{center}
\end{table*}   

\section{Experiments}
In this section, we introduce the developed dataset and compare pre-trained language models~\cite{T5, GPT2} on our dataset \textit{DocBank-TB} to understand the extent to which table caption can be generated automatically. 
Specifically, we compared the table parts to be used,
and retrieval methods to find relevant sentences
by the caption prediction task,
in which,
given the first sentence in a caption,
the prediction accuracy of the remaining sentences was measured.

\subsection{Experimental Settings}

We investigated ten variants of the proposed method, namely, three variants of table information ($T_{r.h}$, $T_{r.o}$, and $T_{r.w}$ represent the use of the row header, others except for the row header, and whole table as input, respectively), and four variants of sentence retrieval methods (Top-N BM25, Author), where  ``Top-N BM25 (N = 1, 2, 3)'' denotes retrieval of Top-N sentences by BM25~\cite{BM25}, ``Author'' denotes extraction of a sentence exactly matched by {\it table indicators} (e.g.: "Table 2:", "Table V")).
In addition, we included a method that does not use any sentences from the paper body as a baseline method. 
Due to our dataset is very small, we did not do fine-tuning to the generative model in this work. We removed the numeral words in the word sequence of the tabular data before concatenating into the generative model, as we observed include tabular numbers without further processing on their semantic meaning or reasoning them led the generative model (especially in the case of GPT2) repetitively generates numbers in the output if the table contains many numeral data (See Table 1). We use the huggingface\footnote{https://huggingface.co} implementation for our experiments. 

\subsection{Evaluation Metrics}
To evaluate the generated text, we used BLEU~\cite{BLEU}, ROUGE-1,2,L~\cite{ROUGE}, and METEOR~\cite{METEOR} to evaluate the generated caption. BLEU evaluates the precision of N-grams in the text are included in the reference text, ROUGE is an evaluation metric evaluates the recall of the N-grams in the reference text are included in the generated text. METEOR is a metric that flexibly measures the match between N-grams by using synonyms and stemming processing, as opposed to the method of evaluating by exact match such as BLEU and ROUGE. These three metrics are widely used for evaluating models that generate text.

\subsection{Evaluation Results}
Table 2 shows the evaluation results in terms of ROUGE, BLEU and METEOR. 
A bold font indicates the highest score for each metric.
Generally, T5 model performed better in BLEU and METEOR, 
while GPT-2 model performed better in ROUGE.
Regarding sentence retrieval methods, 
Top-1 BM25 generally performed better.
This result suggests that using more sentences may increase the risk of retrieving less relevant sentences to the table. 
Whereas, as ``None'' also performed well for some metrics,
there is room for improvement in 
the retrieval methods and 
how to feed retrieved sentences into pre-trained language models.
The evaluation results also show that 
$T_{r.h}$ and $T_{r.w}$ were more effective than $T_{r.o}$,
suggesting that the row header is the most effective and other parts of the table are also helpful for the caption generation task.

\section{Conclusion}

In this paper, we introduced the challenge of table caption generation  for scholarly documents. 
To tackle, we constructed \textit{DocBank-TB}, a dataset consisting of sets of a table and its related information in the scholarly document applicable to our task as well as other natural language tasks. 
We conducted experiments to compare the performance of using different extractive methods to retrieve relevant sentences and three variants of the table-related information from different structures, respectively.  We employed two pre-trained language models to generate a caption and compared different retrieval methods of relevant sentences used as input for caption generation.
We conclude that T5 is the better model for this task, as it outperforms GPT-2 in BLEU and METEOR metrics implying that the generated text are clearer and more precise. Inputting relevant sentences by matching the row header or whole table is effective.
Our comparison experiments and the resources developed in this paper might inspire future research on scholarly documents.
In the future, we wish to extend our research to a large-scale, high-quality table dataset of scholarly documents, also we focus on develop general representation for tabular data that capable to represent the underlying information such as reasoning in numerical data while cooperate with relevant information such as relevant sentences. 
    
\section*{Acknowledgment}
This work was supported by JSPS KAKENHI Grant Numbers JP18H03243 and JP18H03244, and JST PRESTO Grant Number JPMJPR1853, Japan.

\bibliographystyle{IEEEtran}
\bibliography{main}

\end{document}